\pdfoutput=1

\documentclass[11pt]{article}

\usepackage[final]{acl}

\usepackage{times}
\usepackage{latexsym}
\usepackage{amsmath}
\usepackage{booktabs}
\usepackage{graphicx}
\usepackage{multirow} 

\usepackage{algorithm}
\usepackage{algpseudocode}

\usepackage{xcolor}
\usepackage[table]{xcolor}  

\usepackage{amssymb}

\definecolor{customgreen}{rgb}{0.698, 0.906, 0.643}  
\definecolor{customblue}{rgb}{0.867, 0.757, 1}  
\definecolor{lightblue}{RGB}{220,235,250}

\usepackage[T1]{fontenc}

\usepackage[utf8]{inputenc}

\usepackage{microtype}

\usepackage{inconsolata}

\usepackage{graphicx}

\title{Adaptive Planning for Multi-Attribute Controllable Summarization\\with Monte Carlo Tree Search}

 \author{Sangwon Ryu$^{1}$, Heejin Do$^{3,4}$, Yunsu Kim$^{5}$, Gary Geunbae Lee$^{1,2}$, Jungseul Ok$^{1,2}$ \\
  \centering
  \begin{tabular}[t]{c}
    $^{1}$GSAI, POSTECH \quad
    $^{2}$CSE, POSTECH \\
    $^{3}$ETH Zurich \quad
    $^{4}$ETH AI Center \quad
    $^{5}$LILT\\
    \texttt{\{ryusangwon, gblee, jungseul\}@postech.ac.kr}  \\
    \texttt{heejin.do@ai.ethz.ch}\quad\texttt{yunsu.kim@lilt.com}
  \end{tabular}
}

\begin{document}
\maketitle

\begin{abstract}

Controllable summarization moves beyond generic outputs toward human-aligned summaries guided by specified attributes. 
In practice, the interdependence among attributes makes it challenging for language models to satisfy correlated constraints consistently.
Moreover, previous approaches often require per-attribute fine-tuning, limiting flexibility across diverse summary attributes.
In this paper, we propose adaptive \underline{p}lanning for multi-\underline{a}ttribute \underline{co}ntrollable summarization (PACO), a training-free framework that reframes the task as planning the order of sequential attribute control with a customized Monte Carlo Tree Search (MCTS). 
In PACO, nodes represent summaries, and actions correspond to single-attribute adjustments, enabling progressive refinement of only the attributes requiring further control. 
This strategy adaptively discovers optimal control orders, ultimately producing summaries that effectively meet all constraints. 
Extensive experiments across diverse domains and models demonstrate that PACO achieves robust multi-attribute controllability, surpassing both LLM-based self-planning models and fine-tuned baselines. 
Remarkably, PACO with Llama-3.2-1B rivals the controllability of the much larger Llama-3.3-70B baselines. With larger models, PACO achieves superior control performance, outperforming all competitors.

\end{abstract}

\section{Introduction}

Controllable summarization, which tailors summaries to user-specified attributes such as \textit{length},\textit{ extractiveness}, or \textit{topic}, is essential for real-world applications, enabling more personalized outputs. For instance, a student preparing for an exam may prefer a concise summary highlighting only the key topics, whereas a teacher preparing lecture materials may require a more detailed version with high specificity with broad coverage. 

\begin{figure}[t]
\centering
\includegraphics[width=0.48\textwidth]{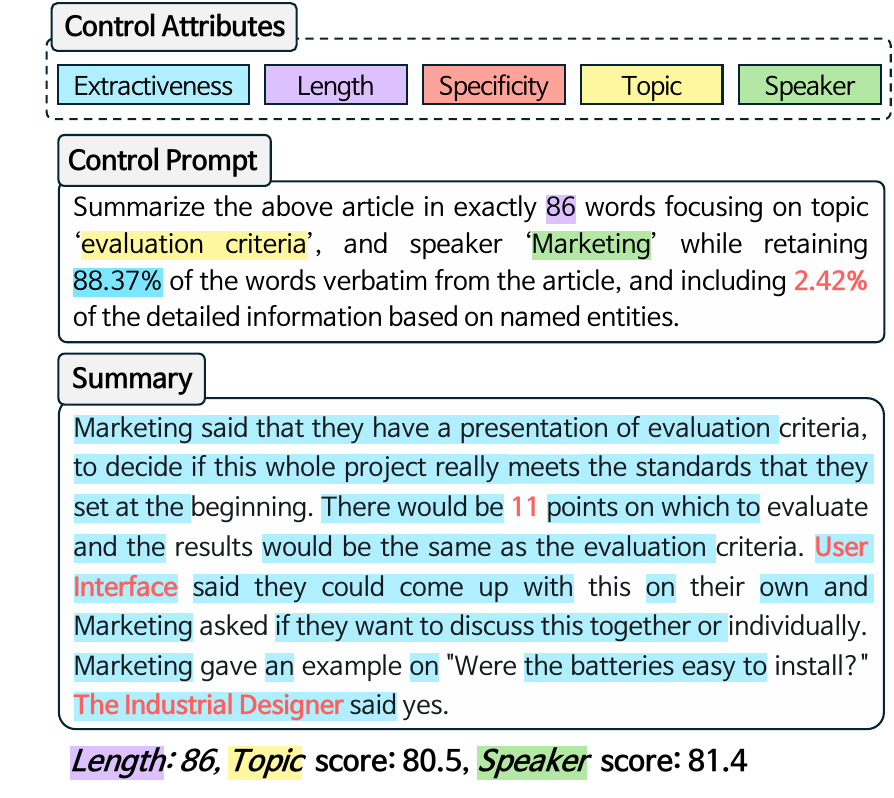}
\caption{Summaries consist of multiple attributes. Our goal is to generate outputs that satisfy diverse user-specified constraints simultaneously.}

\label{fig:example}
\end{figure}

Recent works have explored multi-attribute controllable summarization training with attribute-specific supervision. For instance, \citet{goyal-etal-2022-hydrasum} leveraged a mixture-of-experts (MoE) with each decoder specializing in one attribute, while \citet{zhang-etal-2023-macsum} employed hard prompt tuning (HP) and soft prefix tuning (SP) to train a model for multiple attributes. However, these methods require additional fine-tuning for each attribute, limiting flexibility and generalization to unseen preferences. 
More fundamentally, the autoregressive generation of language models may struggle to enforce multiple correlated constraints simultaneously in a single decoding pass (Figure~\ref{fig:example}).

\begin{figure*}[ht]
\centering
\includegraphics[width=\textwidth]{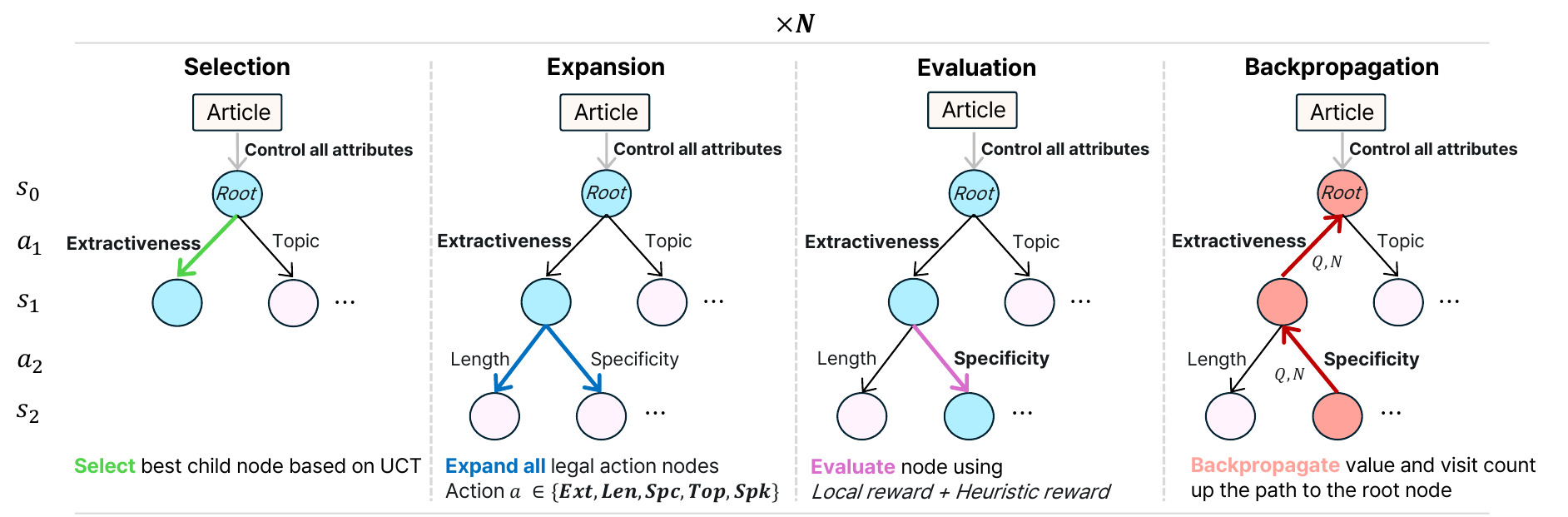}
\caption{Illustration of the MCTS process in PACO. The tree search begins from a summary generated with a prompt that requests control over all attributes, serving as the root node. After all simulations are completed, the node with the highest \textit{degree} is selected from the entire tree during the decision stage.}

\label{fig:main}
\end{figure*}

As multiple attributes often interact in complex ways, achieving full control of all attributes might structurally lead to conflicts; for example, improving extractiveness may inadvertently compromise length control. Moreover, the space of possible attribute control orders grows combinatorially, leaving the unsolved question of how to systematically explore the optimal and effective control paths. 

To address these challenges, we propose adaptive \underline{p}lanning for multi-\underline{a}ttribute \underline{co}ntrollable summarization (PACO), a training-free framework that transforms multi-attribute summarization into a sequential decision-making (i.e., planning) problem. Instead of attempting to enforce all constraints at once, PACO progressively adjusts attributes step by step. 
Specifically, we design a tailored Monte Carlo Tree Search (MCTS) algorithm that explores different control orders at each step by defining nodes at the summary level, while allowing revisiting attributes to adaptively find an optimal control path. 
Since each node encapsulates a complete summary, we can select the summary that maximizes the degree of attribute control once the tree is fully expanded.
To ensure the structured search and evaluation, we categorize attributes by type, distinguishing between \textit{deterministic} attributes, which must match exact user targets, and \textit{non-deterministic} attributes where higher values are preferable. 

We evaluate PACO on diverse domains, including MACSum$_{\text{Doc}}$, MACSum$_{\text{Dial}}$~\cite{zhang-etal-2023-macsum}, and DialogSum~\cite{chen-etal-2021-dialogsum}. In experiments with a range of LLMs, PACO demonstrates robust control performance across models with different sizes and domains. Remarkably, our training-free PACO with a 1B model achieves control performance comparable to that of a 70B baseline model, and PACO with a 70B model outperforms all baselines across all attributes, showing consistently strong controllability. Crucially, PACO achieves these controllability gains without sacrificing summary quality by incrementally adjusting attributes rather than forcing all constraints at once, which may risk compromising summary quality.
Our main contributions are as follows:

\begin{itemize}
\item We introduce PACO, the first framework to cast controllable summarization as a sequential planning problem and adapt MCTS to systematically explore optimal control paths.

\item We define summary-level nodes and categorize attributes by type to assign rewards, thereby enabling flexible and effective enforcement of multiple attribute constraints.

\item Extensive experiments across models and datasets demonstrate PACO's superior controllability and strong alignment with user preferences, even without attribute-specific training.
\end{itemize}

\section{PACO}

Since LLMs struggle to control multiple attributes simultaneously~\cite{ryu-etal-2026-exploring}, we aim to adjust attributes progressively, one at a time. However, sequential controlling of multiple attributes is nontrivial, as the outcome depends on the order of control and the search space of possible orders is combinatorially large. In addition, control attempts may succeed immediately or fail repeatedly, making fixed strategies unreliable and motivating the need for systematic exploration. 

To optimize attribute control planning, we propose PACO, which integrates the MCTS algorithm into multi-attribute controllable summarization. We formulate the attribute control planning process as a Markov Decision Process (MDP). 
Defining nodes at a fine granularity (e.g., token or sentence level) as in prior tree-based approaches with LLMs~\cite{NEURIPS2023_271db992, hao-etal-2023-reasoning, pmlr-v235-wan24c}, can lead to an intractably large search space in long-form generation tasks, such as text summarization. To address this, we define each node at the summary level, reducing search complexity and planning burden on the model. 

\subsection{Problem formulation}

\begin{figure*}[ht]
\centering
\includegraphics[width=\textwidth]{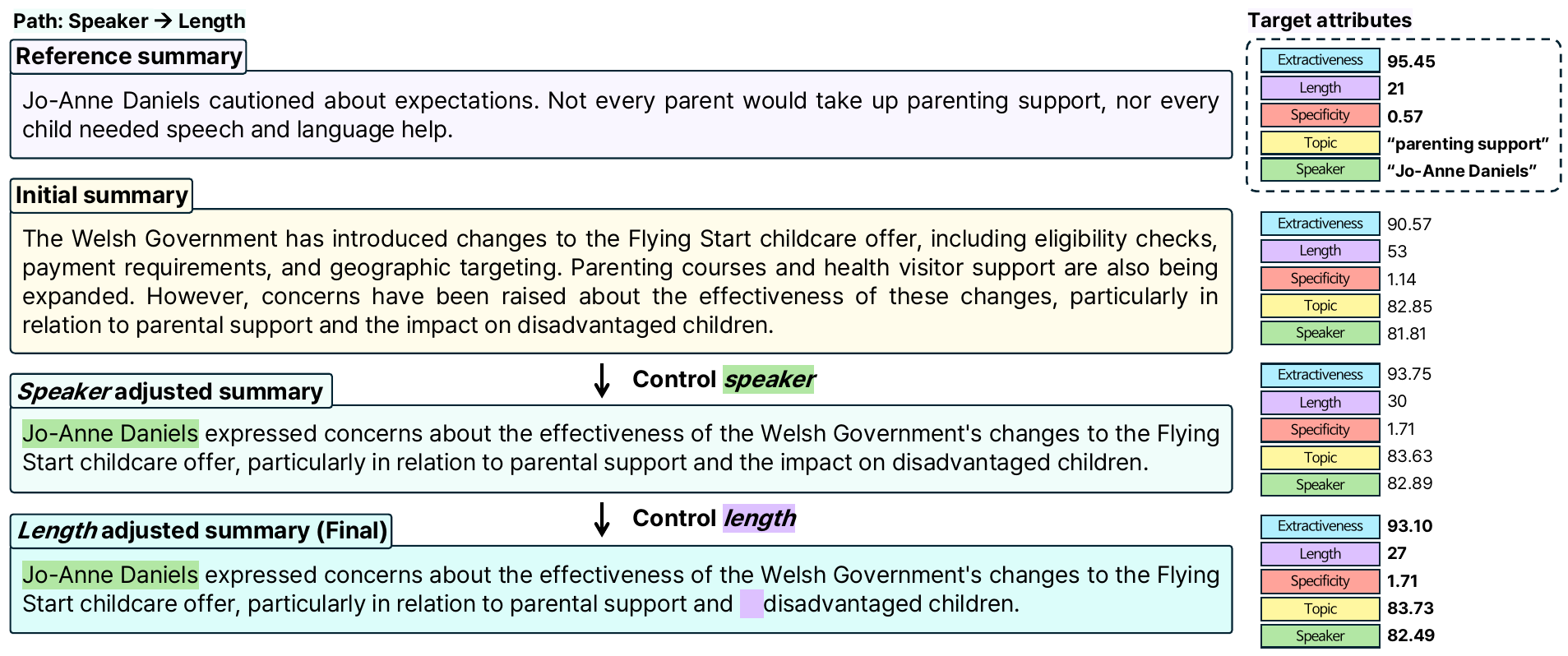}
\caption{An example of PACO adjusting a summary through its planning process. The initial summary shows that LLMs struggle with multiple attribute constraints in a single pass. To address this, PACO successfully refines the summary to meet target attributes. {$\color{customgreen}\blacksquare$} indicates shifts to speaker-focused content; {$\color{customblue}\blacksquare$} highlights removal of unnecessary details to reach the target length. Values beside the reference summary indicate the target attributes, while values beside generated summaries show their measured attribute scores.}
\label{fig:analysis}
\end{figure*}

Starting from an initial summary, PACO identifies under-controlled attributes and progressively adjusts them according to a planned control order, ultimately producing a summary aligned with the target attribute values (Figure~\ref{fig:main}). LLMs serve as the policy $\pi$ and each action $a$ corresponds to controlling a single attribute. We start from an initial summary that reflects all attribute controls, which serves as the root node $s_{0}$, and adaptively search for the optimal attribute control order $[attribute_{1}, attribute_{2}, \ldots, attribute_{n}]$ for each document. Each intermediate summary serves as a state $s$, forming a sequence of transitions from $s_0$ through successive attribute adjustments. At each step $t$, the model determines the action $a_t$ to control a specific attribute and generates the next summary $s_{t+1}$ by taking the full history $s_0, s_1, \ldots, s_t$ as input, enabling informed decisions based on all preceding modifications. We define the tree width $w$ as the number of legal actions, and the tree depth as $d$. The process iterates until reaching a terminal state $T$, which occurs when all attributes have been precisely controlled or when a step limit is exceeded. Figure \ref{fig:analysis} shows an example of how each attribute is adjusted, and we detail the key operations of PACO's customized MCTS algorithm below.

\subsection{MCTS design}

\paragraph{Selection.}

The PACO process begins at the root node $s_{0}$, which is generated by prompting the model to control all attributes in a single initial attempt. The algorithm then explores the search tree by selecting nodes based on a variant of the Predictor Upper Confidence Tree (PUCT)~\cite{10.1007/s10472-011-9258-6} algorithm, using the following equation:

\begin{align}
U(s, a) &= c_{\text{puct}} \cdot \pi_{\theta}(s, a) \cdot \frac{\sqrt{\sum_b N(s, b)}}{1 + N(s, a)} \\
a &= \arg\max_a \left[ Q(s, a) + U(s, a) \right]
\end{align}

Here, $Q(s,a)$ denotes the state-action value and $N(s,a)$ is the visit count for action $a$ at state $s$, both of which are maintained and updated during the search. $N(s,b)$ denotes the visit count of the action $b$ taken from state $s$, where $b$ is one of the possible actions at that state. To balance exploration and exploitation, we use the following term $c_{\text{puct}} = \log\left( \frac{\sum_b N(s, b) + c_{\text{base}} + 1}{c_{\text{base}}} \right) + c_{\text{init}}$, which encourages exploration of less-visited actions while promoting the exploitation of those with high value estimates to maximize expected reward. The selection process continues until a terminal state ($T$) is reached, defined either as a summary that satisfies all attribute constraints or upon reaching a predefined maximum tree depth.

\paragraph{Expansion.}

When a leaf node is reached, we expand it by generating child nodes for all possible actions. The action space is defined as $action \in \{ext, len, spc, top, spk\}$, as each action corresponds to controlling a single attribute. Since the effect of a previously applied action can be altered by subsequent actions, all actions are considered legal throughout the search process.

\paragraph{Evaluation.}

To estimate the value of a node, we use a local reward based on intermediate steps, which captures immediate improvements. The local reward is computed by adapting multi-attribute measurements for controllable summarization~\cite{ryu-etal-2026-exploring}, with each attribute defined as follows:

\begin{itemize}

    \item \textit{Extractiveness}: the proportion of summary words that appear in the source document.
    \item \textit{Length}: the total word count of the summary.
    \item \textit{Specificity}: the ratio of named entities to the total number of words in the summary.
    \item \textit{Topic}: the average embedding similarity $\mathcal{B}$ between $n$ summary words and $k$ topic words: $\frac{1}{k}\sum_{j \in k}\frac{1}{n} \sum_{i \in s}\mathcal{B}(\text{topic}_{j}, \text{word}_{i})$.
    \item \textit{Speaker}: the embedding similarity between the summary and a set of utterances $\mathcal{U}$ from a target speaker in the dialogue, measured by BERTScore$(s, \mathcal{U})$.
\end{itemize}

Using these attribute measurements, we compute the mean absolute deviation (MAD) between the predicted and target values for each requested attribute. We distinguish \textit{deterministic} and \textit{non-deterministic} attributes: \textit{deterministic} attributes, such as \textit{extractiveness}, \textit{length}, and \textit{specificity}, are expected to match user-specified target values, whereas \textit{non-deterministic} attributes, such as \textit{topic} and \textit{speaker}, are evaluated based on their alignment with the target, where higher values indicate better alignments. Therefore, we use the alignment score itself for \textit{non-deterministic} attributes instead of MAD. The total local reward, referred to as the control \textit{degree}, is computed by averaging MAD over \textit{deterministic} attributes ($avg_{\text{det}}$) and adding alignment scores for \textit{non-deterministic} attributes ($avg_{\text{non-det}}$). Since a lower value of $avg_\text{det}$ indicates better performance, we take its reciprocal to align the reward direction. These hyperparameters can be adjusted to control the relative importance of \textit{deterministic} and \textit{non-deterministic} attributes.

\begin{align}
\text{Local reward} = \frac{\alpha}{avg_{\text{det}} + \varepsilon} + \frac{1}{\beta} \cdot avg_{\text{non-det}}
\end{align}

\paragraph{Backpropagation.}

At each simulation's end, we update the visit count and cumulative value estimate $W(s,a)$ for each node along the search path using the simulation result $V(s_l)$ from the leaf node $s_{l}$. The mean action-value $Q(s, a)$ is computed as the cumulative value divided by the visit count. 

\begin{align}
N(s_t, a_t) &\leftarrow N(s_t, a_t) + 1 \\
W(s_t, a_t) &\leftarrow W(s_t, a_t) + V(s_l) \\
Q(s_t, a_t) &= \frac{W(s_t, a_t)}{N(s_t, a_t)}
\end{align}

\paragraph{Decision.}

While node exploration during simulation is guided by stepwise value updates, the final summary is selected based on a fixed \textit{degree}. Unlike standard MCTS approaches that select the most visited or highest-value leaf node~\cite{6145622}, PACO selects the node with the highest degree across the entire tree. This enables PACO to adaptively control a subset of attributes, rather than enforce all of them, allowing for more flexible summarization tailored to each document. See Appendix~\ref{appendix:PACO algorithm} for algorithmic details.

\section{Methods Variants}

\subsection{LLM-Based self-planning}
We examine whether LLMs can perform attribute control planning on their own by introducing two prompt-based self-planning baselines: Implicit and Explicit self-planning.

\paragraph{Implicit self-planning.} We prompt the LLMs to generate the summary in a single pass while implicitly considering which attribute to control first, using \textit{Let's think step-by-step}~\cite{10.5555/3600270.3601883}. The model is encouraged to consider the control order without explicitly generating a separate plan, and to reflect this consideration in the output.

\paragraph{Explicit self-planning.}

We prompt the LLM to generate an explicit control plan from an initial summary and then apply it sequentially. Here, the initial summary is obtained by prompting the model to control all attributes at once. The explicit control plan indicates the sequence to modify the misaligned attributes in the initial summary. Guided by the plan, the model sequentially adjusts each attribute, generating intermediate summaries at each step. Once all attributes in the plan have been controlled, the final summary serves as the output. 
However, we find that unguided explicit self-planning (\textit{base}) typically adjusts each attribute only once or controls unnecessary attributes, often resulting in limited improvement. Since misaligned attributes frequently require multiple rounds of refinement, we introduce an \textit{adaptive} variant that guides the model to selectively target and revisit only the attributes that require further control. Refer to Appendix~\ref{appendix:self-planning} for detailed self-planning prompts.

\subsection{Computation-matched methods}

To verify that PACO’s performance gains do not stem from additional inference-time computation, we design two comparison methods with the same inference-time budget: \textit{Joint-iterative} and \textit{Random sequential} methods. The \textit{Joint-iterative} method jointly controls multiple attributes iteratively, while the \textit{Random sequential} method sequentially adjusts randomly selected attributes. 
Both methods select the best-controlled summary after multiple runs with the same number of generations as PACO, enabling a direct comparison.

\subsection{Heuristic value function}

While traditional MCTS approaches utilize rollouts to estimate the value function~\cite{10.1007/11871842_29, rapid_action_value}, MCTS applications in LLMs typically employ prompt-based heuristic value functions~\cite{NEURIPS2023_271db992, hao-etal-2023-reasoning, yu-etal-2023-prompt} or learned value functions~\cite{pmlr-v235-wan24c, alphamath} to reduce computational cost. Similarly, we design a heuristic value function tailored to controllable summarization. Specifically, we define a heuristic score that reflects the global confidence of the final output, which assesses whether the model can feasibly control all remaining attributes given the current summary and the action path taken so far. Since it is difficult for the model to generate this score as an explicit numeric value, we frame the query as a binary question and use the probability of a "\texttt{Yes}" response as the heuristic score.

\section{Experimental Setup}

\paragraph{Datasets.}

We conduct experiments on two mixed-attribute controllable summarization datasets, MACSum$_{\text{Dial}}$ and MACSum$_{\text{Doc}}$~\cite{zhang-etal-2023-macsum}, and on a topic-focused dialogue summarization dataset, DialogSum~\cite{chen-etal-2021-dialogsum}.  MACSum$_{\text{Dial}}$ is constructed from the QMSum dataset~\cite{zhong-etal-2021-qmsum}, which contains meeting transcripts from three sources: AMI~\cite{ami}, ICSI~\cite{icsi}, and committee meetings from the Welsh Parliament and the Parliament of Canada. MACSum$_{\text{Doc}}$ is based on the CNN/DailyMail~\cite{see-etal-2017-get}, a news domain dataset. DialogSum consists of real-life scenario that covers general topics from everyday life. Notably, only MACSum$_{\text{Dial}}$ includes the \textit{speaker}.

\paragraph{Models.}

We demonstrate the robustness of our approach by applying it to various LLMs of different sizes, including the Llama series (Llama-3.2-1B-Instruct and Llama-3.3-70B-Instruct)~\cite{touvron2023llamaopenefficientfoundation, grattafiori2024llama3herdmodels} and Qwen2.5-7B-Instruct~\cite{qwen1, qwen2.5}. As baselines, we compare against LLM-based self-planning methods, namely implicit self-planning and explicit self-planning (including both \textit{base} and \textit{adaptive} versions), as well as hard prompt tuning combined with soft prefix tuning (HP+SP)~\cite{10.5555/3455716.3455856, li-liang-2021-prefix}, reimplemented following \citet{zhang-etal-2023-macsum} based on BART$_{\text{large}}$~\cite{lewis-etal-2020-bart}. We measure embedding similarity using BERTScore~\cite{zhang2020bertscore} and extract named entities with FLAIR~\cite{akbik2019flair}, a well-established named entity recognition (NER) model trained on OntoNotes 5~\cite{pradhan-etal-2013-towards}, which covers various domains, including news and conversational speech. 

\paragraph{Metrics.}

We adopt different evaluation for \textit{deterministic} and \textit{non-deterministic} attributes. While we compute the mean absolute deviation (MAD) between the target and generated attribute values for \textit{deterministic} attributes (lower is better), we directly evaluate the generated values for \textit{non-deterministic} attributes (higher is better). Although our main goal is to achieve controllable summarization, maintaining overall summary quality remains important. To this end, we additionally evaluate the quality of the generated summaries using ROUGE-1~\cite{lin-2004-rouge} and BERTScore F1~\cite{zhang2020bertscore}.

\begin{table*}[ht]
\centering
\scalebox{0.78}
{\begin{tabular}{l|c|ccccc|cc}
\toprule
\textbf{Model} & \textbf{\# of Params} & \textbf{Ext ($\downarrow$)} & \textbf{Len ($\downarrow$)} & \textbf{Spc ($\downarrow$)} & \textbf{Top ($\uparrow$)} & \textbf{Spk ($\uparrow$)} & \textbf{ROUGE ($\uparrow$)} & \textbf{BERTScore ($\uparrow$)}\\
\midrule
Reference summary & - & 0.00 & 0.00 & 0.00 & 0.796 & 0.802 & - & -\\
\midrule
HP+SP (BART$_{\text{large}}$)$^{*}$ & 406M & 6.66 & 34.66 & 7.08 & 0.807 & 0.804 & 0.315 & 0.871 \\
\midrule
Llama-3.2-Instruct & 1B & 10.79 & 55.68 & 9.30 & 0.783 & \textbf{0.795} & 0.270 & 0.854 \\
\rowcolor{lightblue}
\quad$\circ$ PACO & 1B & \textbf{9.30} & \textbf{17.96} & \textbf{7.22} & \textbf{0.792} & 0.794 & 0.288 & 0.859 \\
\midrule
Qwen2.5-Instruct & 7B & 9.70 & 17.82 & 6.99 & 0.797 & \textbf{0.795} & 0.301 & 0.867 \\
\rowcolor{lightblue}
\quad$\circ$ PACO & 7B & \textbf{8.72} & \textbf{11.79} & \textbf{5.43} & \textbf{0.799} & 0.794 & 0.302 & 0.868 \\
\midrule
Llama-3.3-Instruct & 70B & 6.43 & 15.72 & 7.11 & 0.800 & \textbf{0.798} & 0.328 & 0.871 \\
\quad$\circ$ Implicit self-planning & 70B & 7.35 & 27.70 & 8.09 & 0.802 & 0.795 & 0.304 & 0.869 \\
\quad$\circ$ Explicit self-planning & 70B & 7.44 & 28.19 & 7.32 & \textbf{0.808} & 0.794 & 0.287 & 0.869 \\
\quad$\circ$ Explicit self-planning+ & 70B & 7.08 & 24.52 & 7.32 & 0.801 & 0.795 & 0.312 & 0.869 \\
\quad$\circ$ Joint-iterative & 70B & 5.19 & 11.19 & 5.18 & 0.797 & 0.797 & 0.319 & 0.867 \\
\quad$\circ$ Random sequential & 70B & 5.44 & 11.16 & 4.24 & 0.797 & 0.797 & 0.322 & 0.869 \\
\rowcolor{lightblue}\quad$\circ$ PACO  & 70B & \textbf{4.91} & \textbf{7.63} & \textbf{3.81} & 0.795 & \textbf{0.798} & 0.328 & 0.869 \\
\bottomrule
\end{tabular}}
\caption{Controllability evaluation results on MACSum$_{\text{Dial}}$. {Explicit self-planning} denotes the \textit{base} version, while {Explicit self-planning+} refers to the \textit{adaptive} variant. \textbf{Bold} indicates the best controllability within the same baseline model; $^{*}$ marks models trained on the corresponding data; ↑ indicates higher is better, and ↓ indicates lower is better.}

\label{tab:main_dial}
\end{table*}

\begin{table*}[ht]
\centering
\scalebox{0.8}
{\begin{tabular}{l|c|cccc|cc}
\toprule
\textbf{Model} & \textbf{\# of Params} & \textbf{Ext ($\downarrow$)} & \textbf{Len ($\downarrow$)} & \textbf{Spc ($\downarrow$)} & \textbf{Top ($\uparrow$)} & \textbf{ROUGE ($\uparrow$)} & \textbf{BERTScore ($\uparrow$)} \\
\midrule
Reference summary & - & 0.00 & 0.00 & 0.00 & 0.806 & - & - \\
\midrule
HP+SP (BART$_{\text{large}}$)$^{*}$ & 406M & 9.04 & 19.43 & 4.55 & 0.803 & 0.327 & 0.881 \\
\midrule
Llama-3.2-Instruct & 1B & 7.84 & 9.67 & 4.27 & 0.792 & 0.330 & 0.879 \\
\rowcolor{lightblue}
\quad$\circ$ PACO & 1B & \textbf{7.53} & \textbf{4.58} & \textbf{3.73} & \textbf{0.796} & 0.326 & 0.879 \\
\midrule
Qwen2.5-Instruct & 7B & 7.39 & 13.78 & 5.42 & 0.793 & 0.321 & 0.879 \\
\rowcolor{lightblue}
\quad$\circ$ PACO & 7B & \textbf{6.37} & \textbf{7.03} & \textbf{4.18} & \textbf{0.797} & 0.319 & 0.880 \\
\midrule
Llama-3.3-Instruct & 70B & 8.03 & 10.91 & 3.58 & 0.802 & 0.308 & 0.878 \\
\quad$\circ$ Implicit self-planning & 70B & 8.14 & 17.05 & 4.56 & 0.803 & 0.296 & 0.875 \\
\quad$\circ$ Explicit self-planning & 70B & 8.22 & 15.64 & 3.89 & \textbf{0.808} & 0.285 & 0.875 \\
\quad$\circ$ Explicit self-planning+ & 70B & 8.19 & 13.86 & 3.99 & 0.800 & 0.286 & 0.873 \\
\quad$\circ$ Joint-iterative & 70B & 4.58 & 7.15 & 3.35 & 0.798 & 0.298 & 0.873  \\
\quad$\circ$ Random sequential & 70B & 5.17 & 6.62 & 3.19 & 0.797 & 0.309 & 0.875 \\
\rowcolor{lightblue}
\quad$\circ$ PACO & 70B & \textbf{4.49} & \textbf{4.81} & \textbf{2.84} & 0.794 & 0.322 & 0.876 \\
\bottomrule
\end{tabular}}
\caption{Controllability evaluation results on MACSum$_{\text{Doc}}$, which does not include the \textit{speaker} attribute.}
\label{tab:main_doc}

\end{table*}

\begin{table}[t]
\centering
\scalebox{0.65}
{\begin{tabular}{l|c|cccc}
\toprule
\textbf{Model} & \textbf{Params} & \textbf{Ext ($\downarrow$)} & \textbf{Len ($\downarrow$)} & \textbf{Spc ($\downarrow$)} & \textbf{Top ($\uparrow$)} \\
\midrule
Reference summary & - & 0.00 & 0.00 & 0.00 & 0.817 \\
\midrule
Llama-3.2-Instruct & 1B & 20.45 & 15.65 & 52.43 & 0.815  \\
\rowcolor{lightblue}
\quad$\circ$ PACO & 1B & \textbf{14.17} & \textbf{6.28} & \textbf{28.48} & \textbf{0.825} \\
\midrule
Qwen2.5-Instruct & 7B & 12.08 & 5.20 & 26.62 & 0.817  \\
\rowcolor{lightblue}
\quad$\circ$ PACO & 7B & \textbf{8.71} & \textbf{3.30} & \textbf{19.14} & \textbf{0.820}  \\
\midrule
Llama-3.3-Instruct & 70B & 14.91 & 2.26 & 20.82 & \textbf{0.829}  \\
\rowcolor{lightblue}
\quad$\circ$ PACO  & 70B & \textbf{8.35} & \textbf{1.56} & \textbf{10.20} & 0.828  \\
\bottomrule
\end{tabular}}
\caption{Evaluation results on DialogSum. While annotator-specific attributes lead to varying control trends, PACO consistently outperforms all baselines.}

\label{tab:main_dialogsum}
\end{table}

\section{Main Results}

\paragraph{Controllability results.}

While target attributes can be arbitrarily chosen, we use the values of the reference summaries to enable direct comparison. For \textit{topic} and \textit{speaker}, we use values provided in the dataset. As shown in Table~\ref{tab:main_dial}, smaller-scale LLM baselines struggle to control attributes, particularly \textit{length}, resulting in excessively high MAD on MACSum$_{\text{Dial}}$. Since this dataset contains long and complex meeting transcripts, even large models such as Llama-3.3-70B struggle to control \textit{length}, with MAD exceeding 15.
In contrast, PACO consistently shows strong attribute control across different models. Notably, it reduces the MAD for \textit{length} from 55.68 to 17.96 on the 1B model, which is comparable to the performance of the 70B baseline. 
On Llama-3.3-70B, PACO achieves an average MAD of approximately 5 over \textit{deterministic} attributes, demonstrating precise control and clearly outperforming all baselines. Applying PACO to Qwen2.5-7B also yields substantial improvements over the base model, with performance falling between the 1B and 70B Llama models, highlighting its generalizability.
Importantly, PACO outperforms both budget-matched methods, indicating that its improvements arise from structured planning rather than additional attempts. In addition, \textit{Random sequential} surpasses \textit{Joint-iterative}, supporting the effectiveness of our step-wise control framework.

\paragraph{Robustness across datasets.}
As shown in Table~\ref{tab:main_doc}, PACO again significantly outperforms all baselines on MACSum$_{\text{Doc}}$. Remarkably, the 1B PACO model even surpasses the 70B baseline, while our 70B model exhibits dominant controllability, clearly surpassing all other models. Compared to MACSum$_{\text{Dial}}$, which consists of longer and more complex input texts, all models shows better controllability on MACSum$_{\text{Doc}}$ with simpler inputs.
These results highlight that PACO maintains robust control across domains and input complexities, whereas baseline methods show a notable decline as inputs become longer and more complex.

We further evaluate PACO on DialogSum. Table~\ref{tab:main_dialogsum} shows that PACO achieves substantial gains in controllability across all model sizes. Interestingly, the controllability pattern on DialogSum differs from that of the MACSum datasets. While \textit{length} is the most difficult and \textit{specificity} the easiest to control in MACSum, the reverse holds for DialogSum. This discrepancy may stem from domain-specific properties or variation in annotation style, as human-written summaries can differ across annotators. These results underscore the effectiveness of adaptive control in PACO, which flexibly adjusts to the unique characteristics of each dataset.

\begin{figure}[t]
\centering
\includegraphics[width=0.46\textwidth]{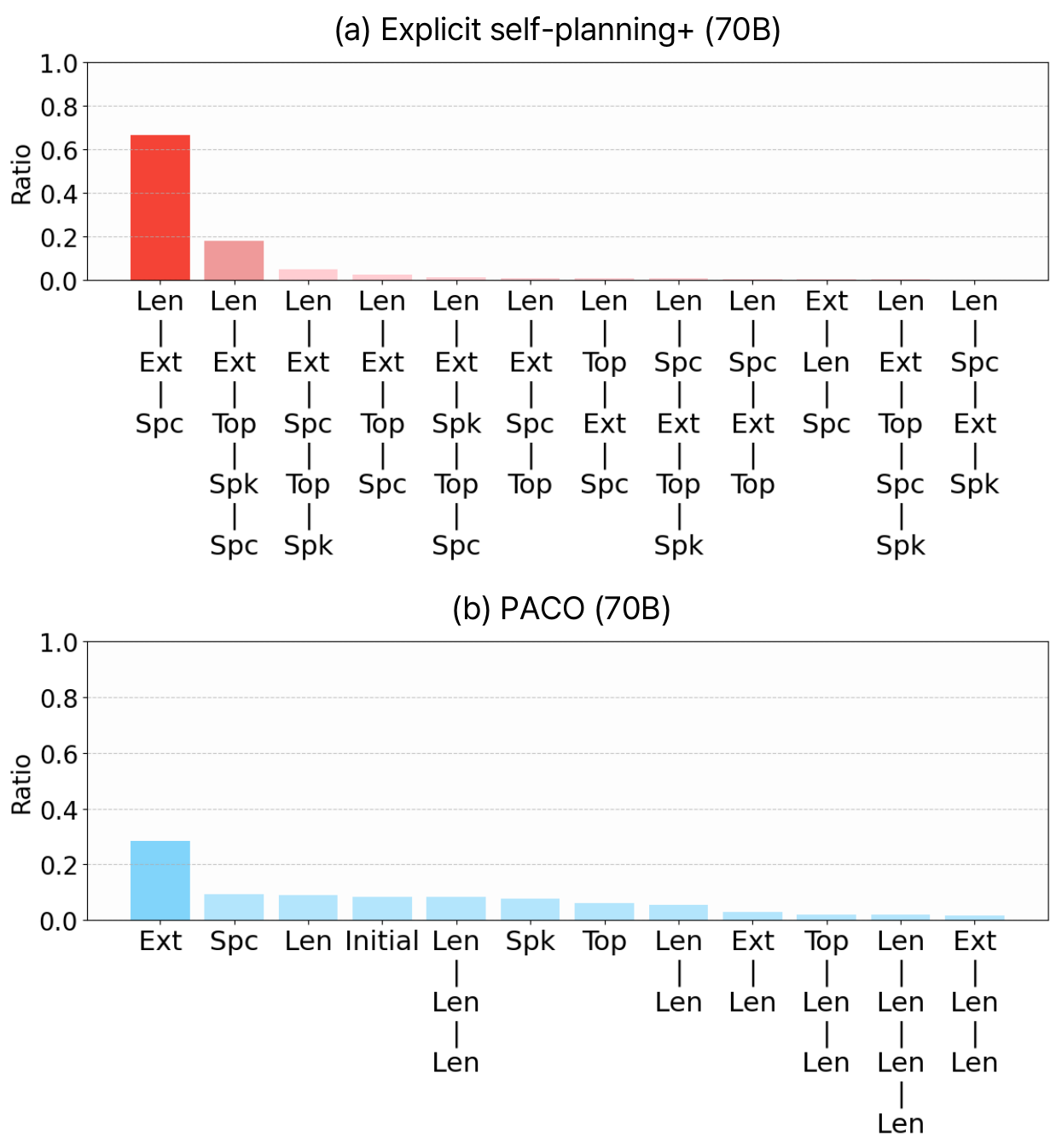}
\caption{(a) LLMs often over-control and lack diversity in their self-generated plans, whereas (b) PACO controls only the necessary attributes for each instance. We visualize the top 10 plans for each method.}
\label{fig:plan}
\end{figure}

\paragraph{Balancing between attribute types.}
The results show that LLM-based models are more effective at controlling \textit{non-deterministic} than \textit{deterministic} attributes, with \textit{topic} and \textit{speaker} scores comparable to the reference summaries. Given that attribute types can be prioritized, we place higher weight on the \textit{deterministic} attributes. We provide more detailed experiments on balancing control performance across attribute types in Appendix~\ref{appendix:balancing}. Although HP+SP is explicitly trained for attribute control, it often fails to follow instructions. We consider this to be due to structural constraints of the encoder–decoder architecture and to the use of vague supervision for \textit{deterministic} attributes (e.g.,"high" rather than precise targets). 

\begin{figure}[t]
\centering
\includegraphics[width=0.48\textwidth]{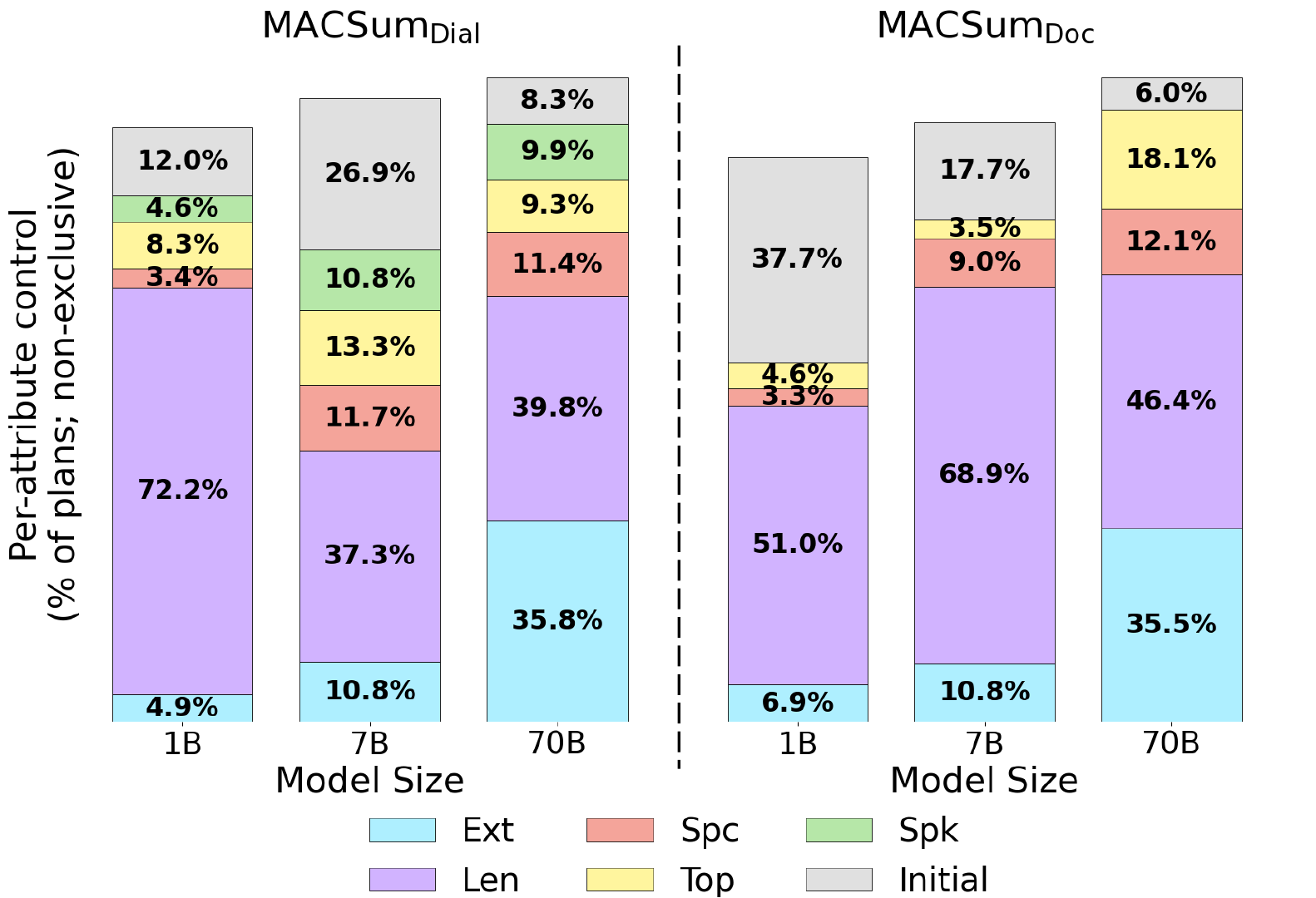}
\caption{Each bar shows the frequency of attribute control per model size, with repeated attributes in a plan counted once. Percentages indicate relative proportions. \textit{Initial} refers to the state where all attributes were controlled simultaneously at the beginning.}
\label{fig:attribute ratio}
\end{figure}

\paragraph{Comparison with self-planning.}

We evaluate whether LLMs can perform attribute control planning (Tables~\ref{tab:main_dial},~\ref{tab:main_doc}). The results show that both Implicit and Explicit self-planning fail to generate effective plans, performing even worse than the baseline. In particular, Implicit self-planning exhibits the weakest control performance. The \textit{adaptive} version, Explicit self-planning+, incorporates soft constraints into the prompt and improves on the base version; however, it still lags behind the baseline. These findings highlight that LLMs struggle with attribute planning in multi-attribute controllable, underscoring the need for a more effective planning strategy to guide the generation process.

As shown in Figure~\ref{fig:plan}, PACO selectively adjusts only the necessary attributes starting from the initial summary, resulting in diverse and well-balanced control plan distributions. In contrast, Explicit self-planning+, despite being prompted to plan only for necessary adjustments, tends to produce repetitive and imbalanced plans across most data points. This highlights that LLMs struggle with planning in controllable summarization.

\begin{table*}[ht]
\centering
\scalebox{0.8}
{\begin{tabular}{l|c|ccccc}
\toprule
\textbf{Model} & \textbf{\# of Params} & \textbf{Extractiveness ($\downarrow$)} & \textbf{Length ($\downarrow$)} & \textbf{Specificity ($\downarrow$)} & \textbf{Topic ($\uparrow$)} & \textbf{Speaker ($\uparrow$)} \\
\midrule
PACO (L) & 1B & \textbf{9.30} & 17.96 & \textbf{7.22} & 0.792 & \textbf{0.794} \\
PACO (H) & 1B & 9.56 & 21.98 & 7.97 & 0.791 & 0.793 \\
PACO (L + H) & 1B & 9.56 & \textbf{16.12} & 7.63 & \textbf{0.793} & \textbf{0.794} \\
\midrule
PACO (L) & 70B & \textbf{4.91} & 7.63 & \textbf{3.81} & 0.795 & \textbf{0.798} \\
PACO (H) & 70B & 5.06 & 7.59 & 3.94 & \textbf{0.796} & \textbf{0.798} \\
PACO (L + H) & 70B & 5.16 & \textbf{7.56} & 4.28 & 0.795 & 0.796 \\
\bottomrule
\end{tabular}}
\caption{Ablation study of the value function. `L' denotes the local reward, and `H' denotes the heuristic score.}

\label{tab:ablation_value_function}
\end{table*}

\paragraph{Quality evaluation.}
Focusing too heavily on attribute control may risk degrading the summary quality, so we also evaluate the overall summary quality (Table~\ref{tab:main_dial},~\ref{tab:main_doc}). 
Notably, by incrementally controlling attributes rather than enforcing all constraints simultaneously, PACO avoids potential quality degradation and preserves summary quality comparable to the baseline. Although LLMs have already demonstrated strong summarization capabilities~\cite{goyal2023news, pu2023summarization, zhang-etal-2024-benchmarking, ryu24_interspeech}, PACO not only excels in control performance but also preserves their high generation quality. Additionally, although LLMs tend to generate more paraphrased outputs, which often lead to lower ROUGE scores compared to trained encoder-decoder models, they can still achieve higher ROUGE scores when given precise control instructions.

\begin{figure}[t]
\centering
\includegraphics[width=0.48\textwidth]{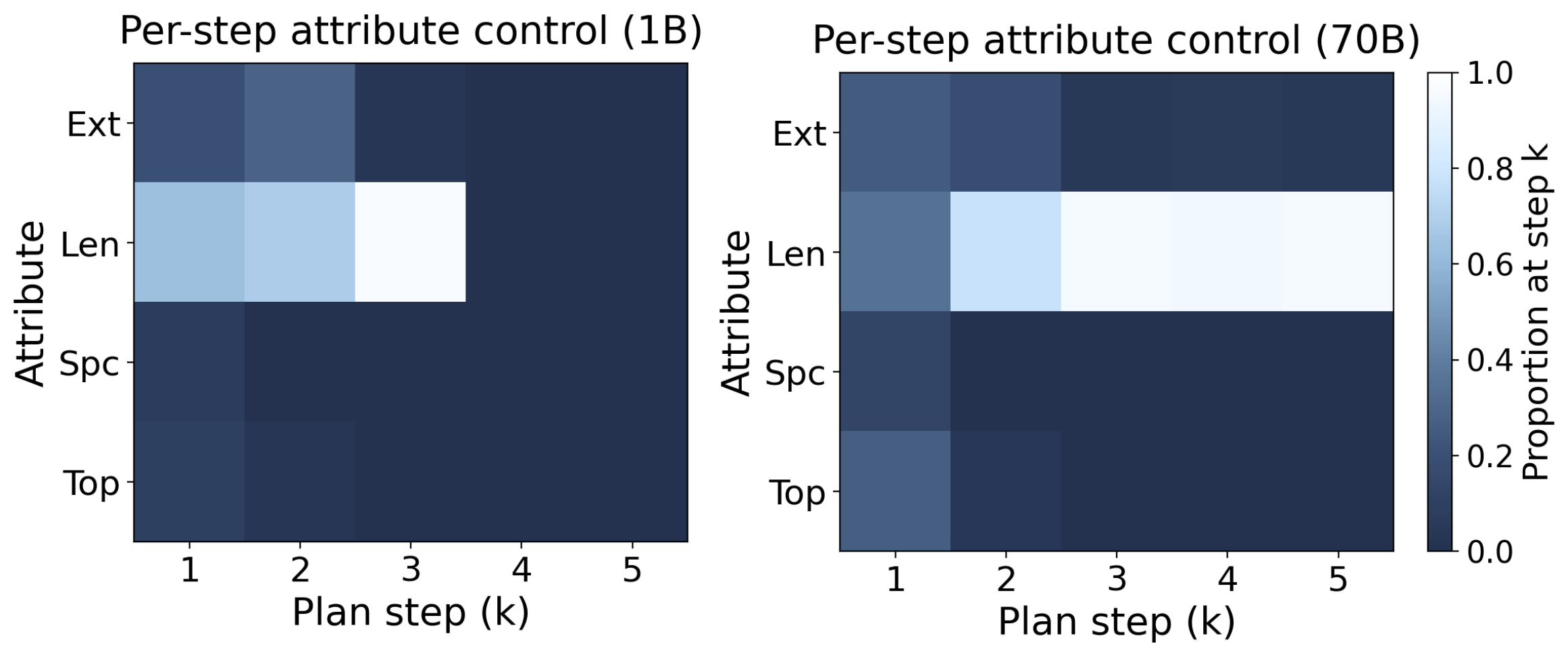}
\caption{Average attribute controlled at each step in DialogSum. Later steps mostly control \textit{length}, which may be due to high correlation with other attributes.}
\label{fig:step}
\end{figure}

\section{Analysis}

\paragraph{Frequency of attribute control.}

In Figure~\ref{fig:attribute ratio}, we analyze the attributes adjusted by PACO across model sizes and domains. 
As the model size increases, fewer initial summary is selected. This suggests that larger models adjust attributes more effectively. Particularly, they more often control \textit{extractiveness} and \textit{specificity}, demonstrating their capacity for sophisticated control.
Across all model sizes, \textit{length} is the most frequently adjusted attribute, likely due to its strong correlation with others and the resulting need for additional revisions.
In MACSum$_{\text{Dial}}$, which has longer and more complex inputs, the initial summary is selected less in total than the MACSum$_{\text{Doc}}$.
This indicates that longer inputs often require additional adjustments to satisfy multiple constraints.

\paragraph{Step-wise control patterns.}

We analyze the average attribute controlled by PACO at each step across variable-length plans (Figure~\ref{fig:step}).
The 70B model adjusts diverse attributes in the early steps, while later steps tend to focus on \textit{length}, as it can be highly affected by other attributes. 
In contrast, the 1B model shows limited controllability gains from deeper adjustments and rarely attempts to control attributes beyond \textit{length}.

\paragraph{Ablation study on the value function.}

We present an ablation study comparing different strategies for computing node values, comparing control degree at the current step with heuristic scoring, a common choice in LLM-based MCTS (Table~\ref{tab:ablation_value_function}). The results show that the heuristic score provides little benefit as a value function, especially for the 1B model. 
Combining both signals offers only marginal gains, which are insufficient to offset the additional cost.
Thus, using only the local reward is both more efficient and effective, likely because predicting whether a partially controlled summary will meet all remaining attributes is nontrivial.

\section{Related Work}

\paragraph{Controllable summarization.}

Prior work on controllable summarization has focused primarily on single-attribute control~\cite{zhong-etal-2021-qmsum, liu-chen-2021-controllable, dou-etal-2021-gsum, chan-etal-2021-controllable, mao-etal-2022-dyle, zhang-etal-2022-summn, bahrainian-etal-2022-newts, ahuja-etal-2022-aspectnews, liu-etal-2022-length, maddela-etal-2022-entsum, mehra-etal-2023-entsumv2, xu-etal-2023-lmgqs, pagnoni-etal-2023-socratic, wang-etal-2023-instructive, retkowski-waibel-2025-zero, gu-etal-2025-length, ryu-etal-2026-exploring}, most commonly targeting adjustments to \textit{length} and \textit{topic}~\cite{urlana-etal-2024-controllable}. To control diverse attributes,~\citet{he-etal-2022-ctrlsum} introduced a framework capable of controlling various attributes, such as \textit{entity}, \textit{length}, and \textit{purpose}, though these were not controlled simultaneously.

Recently, there has been growing interest in controlling multiple attributes simultaneously~\cite{fan-etal-2018-controllable, goyal-etal-2022-hydrasum, zhang-etal-2023-macsum}. \citet{zhang-etal-2023-macsum} introduced mixed-attribute controllable summarization, while \citet{goyal-etal-2022-hydrasum} leveraged MoE to control multiple attributes jointly.
However, these methods require additional training for each attribute, making them impractical when the number of attributes increases. In contrast, PACO leverages LLMs without any attribute-specific training, using planning via MCTS to discover optimal control paths and enable simultaneous control of all target attributes.

\paragraph{Tree search for LLMs.}

Tree search has predominantly been applied to reasoning tasks, where problems are decomposed into sub-questions and represented as nodes in a search tree to facilitate step-by-step reasoning toward the correct answer~\cite{NEURIPS2023_271db992, hao-etal-2023-reasoning, pmlr-v235-wan24c, alphamath, zhang2024accessinggpt4levelmathematical, xie2024montecarlotreesearch, lee-etal-2025-semantic}. \citet{NEURIPS2023_271db992} frame each node as a partial solution and search the tree to solve complex problems. \citet{hao-etal-2023-reasoning} treats the language model as a world model, defining task-specific states and actions. Whereas prior work has applied MCTS to identify reasoning paths that lead to a correct final answer, our task requires that the entire decoding process satisfy multiple constraints. To address this, we tailor MCTS to the controllable summarization setting, defining each node at the summary level rather than at the token or sentence level. In addition, since the degree of control over a previously adjusted attribute may change as subsequent actions are taken, we allow the same attribute to be adjusted multiple times during the search.

\section{Conclusion}

We propose PACO, an adaptive planning method that incorporates Monte Carlo Tree Search into multi-attribute controllable summarization, enabling effective control over multiple attributes. Since it is challenging for language models to enforce all constraints simultaneously in a single pass, PACO incrementally adjusts attributes by constructing optimal control paths, revising only the necessary attributes. As a result, it demonstrates robust and consistent control across various models and domains while preserving summary quality.

\section*{Limitations}

Although PACO offers strong controllability across multiple attributes, we note a few practical limitations and directions for extension. First, while summary-level nodes help reduce the search space, the tree search remains computationally expensive (refer to Appendix~\ref{appendix:computational cost}). Finding an optimal control path requires deeper simulations, leading to longer runtime. Nevertheless, PACO operates entirely at test time and requires no additional training, it is practically valuable despite the added computational overhead. To mitigate this limitation, future work could explore more efficient search-time heuristics or incorporate approximation strategies to reduce computational cost without sacrificing control quality. Second, incorporating the optimization of quality dimensions, as explored in previous work such as \citet{ryu-etal-2024-multi} and \citet{song-etal-2025-learning}, could expand controllability beyond attribute alignment to broader quality dimensions such as \textit{coherence}, \textit{consistency}, \textit{relevance}, and \textit{fluency}. To support more comprehensive and user-tailored summarization, future work could extend PACO to accommodate a broader range of attribute types.

\section*{Ethical Statement}

This paper focuses on applications in controllable summarization and raises no ethical concerns. All datasets used are publicly available, and AI assistance was employed exclusively for grammar correction.

\section*{Acknowledgments}

This work was supported by the National Research Foundation of Korea (NRF) grant funded by the Korea government (MSIT) (No. RS-2023-00217286) (45\%); by Smart HealthCare for Police Officers Program (www.kipot.or.kr) through the Korea Institutes of Police Technology (KIPoT) funded by the Korean National Police Agency (KNPA, Korea) (No. RS-2022-PT000186) (45\%); and by Institute of Information \& communications Technology Planning \& Evaluation (IITP) grant funded by the Korea government (MSIT) (No.RS-2019-II191906, Artificial Intelligence Graduate School Program (POSTECH)) (10\%).

\bibliography{main}

\clearpage

\appendix

\section{PACO Algorithm}
\label{appendix:PACO algorithm}

Algorithm~\ref{algo:PACO} outlines the PACO procedure. The algorithm includes selection, expansion, evaluation, and backpropagation during the simulation phase, followed by a decision step that selects the final summary from the entire tree.

\begin{algorithm}
\caption{PACO ($\mathcal{M}_\theta$)}\label{algo:PACO}
\begin{algorithmic}[1]
\Require LM $\mathcal{M}_\theta$, attribute measure $f$
\Require article $x$, target attributes $a^*$
\Require controllable attributes $\mathcal{A}$, hyperparameters: simulations $n$, max depth $d$

\State \emph{\textcolor{gray}{// initialize root node with summary controlling all attributes}}
\State $y_0 \gets \mathcal{M}_\theta(x, a^*, \mathcal{A})$
\State $\hat{a}_0 \gets f(x, y_0)$
\State $deg_0 \gets \text{degree}(\hat{a}_0, a^*)$
\State $s_0 \gets \text{node}(y_0, \hat{a}_0, deg_0, \text{depth}=0)$

\For{$i = 1$ to $n$}
    \State initialize state $s \gets s_0$

    \State \emph{\textcolor{gray}{// selection}}
    \While{$s$ is not a leaf and not terminated \textbf{and} $\text{depth}(s) < d$}
        \State $a \gets \arg\max_{a' \in \mathcal{A}} \mathrm{PUCT}(s, a')$
        \State $s \gets \text{child}(s, a)$
    \EndWhile

    \State \emph{\textcolor{gray}{// expansion}}
    \If{$s$ is a leaf and not terminated}
        \State create child nodes $\{ \text{child}(s, a') \}_{a' \in \mathcal{A}}$
    \EndIf

    \State \emph{\textcolor{gray}{// evaluation}}
    \State $a' \gets \arg\max_{a' \in \mathcal{A}} \mathrm{PUCT}(s, a')$
    \State $s' \gets \text{child}(s, a')$
    
    \If{$s'$ has no summary}
        \State $y \gets \mathcal{M}_\theta(x, a^*, a', \text{history}(s'))$
        \State $\hat{a} \gets f(x, y)$
        \State $deg \gets \text{degree}(\hat{a}, a^*)$
        \State store $y$, $\hat{a}$, $deg$ in $s'$
    \Else
        \State retrieve $y$, $\hat{a}$, $deg$ from $s'$
    \EndIf

    \State \emph{\textcolor{gray}{// backpropagation}}
    \While{$s' \neq s_0$}
        \State update stats at $s'$
        \State $s' \gets$ parent of $s'$
    \EndWhile
\EndFor

\State \emph{\textcolor{gray}{// decision}}
\State $s^* \gets \arg\max_{s \in \text{Tree}} \text{degree}(s)$
\State \Return summary $y^*$ from $s^*$
\end{algorithmic}
\end{algorithm}

\begin{table*}[ht]
\centering
\scalebox{0.78}
{\begin{tabular}{l|c|ccccc}
\toprule
\textbf{Model} & \textbf{\# of Params} &  \textbf{Extractiveness ($\downarrow$)} & \textbf{Length ($\downarrow$)} & \textbf{Specificity ($\downarrow$)} & \textbf{Topic ($\uparrow$)} & \textbf{Speaker ($\uparrow$)} \\
\midrule
Reference summary & - & 0.00 & 0.00 & 0.00 & 0.796 & 0.802 \\
\midrule
Qwen2.5-Instruct & 7B & 9.70 & 17.82 & 6.99 & 0.797 & 0.795 \\
\quad$\circ$ PACO ($\beta=10$)  & 7B  & \textbf{8.72} & \textbf{11.79} & \textbf{5.43} & 0.799 & 0.794 \\
\quad$\circ$ PACO ($\beta=0.5$) & 7B  & 8.98 & 12.66 & 5.96 & 0.799 & 0.794 \\
\quad$\circ$ PACO ($\beta=0.2$) & 7B   & 8.85 & 13.22 & 5.99 & 0.800 & 0.794 \\
\quad$\circ$ PACO ($\beta=0.1$) & 7B   & 8.88 & 13.45 & 5.99 & 0.800 & 0.794 \\
\quad$\circ$ PACO ($\beta=0.01$) & 7B   & 9.31 & 17.94 & 6.44 & \textbf{0.801} & \textbf{0.795} \\
\midrule
Llama-3.3-Instruct & 70B  & 6.43 & 15.72 & 7.11 & 0.800 & 0.798 \\
\quad$\circ$ PACO ($\beta=10$) & 70B  & 4.91 & \textbf{7.63} & \textbf{3.81} & 0.795 & 0.798 \\
\quad$\circ$ PACO ($\beta=0.5$)  & 70B  & \textbf{4.83}  & 8.48 & 4.34 & 0.796 & 0.799 \\
\quad$\circ$ PACO ($\beta=0.2$) & 70B  & 4.90 & 11.01 & 4.81 & 0.798 & 0.800 \\
\quad$\circ$ PACO ($\beta=0.1$) & 70B  &  5.04 & 12.41 & 5.39 & 0.799  & 0.800 \\
\quad$\circ$ PACO ($\beta=0.01$) & 70B  & 5.88  &  15.77 & 5.80 & \textbf{0.800} &  \textbf{0.801} \\
\bottomrule
\end{tabular}}
\caption{Effect of weighting attribute types. Adjusting the weights shifts the emphasis between \textit{deterministic} and \textit{non-deterministic} attributes. As $\beta$ decreases, more weight is placed on \textit{non-deterministic} attributes, resulting in improved scores for \textit{topic} and \textit{speaker}.}\label{tab:beta}
\end{table*}

\section{Hyperparameters}
\label{appendix:hyperparameters}
We set the maximum tree depth $d$ to 5 and perform 8 simulations per search. To compute the local reward, we use $\alpha$ = 1 for \textit{deterministic} attributes and $\beta$ = 10 for \textit{non-deterministic} attributes, balancing their scales. All budget-matched methods use the same number of generations as PACO’s simulation count. We adopt most MCTS hyperparameters from~\citet{Silver2017} and~\citet{Schrittwieser2020}, including $c_{\text{base}}$ = 19652 and $c_{\text{init}}$ = 1.25. we follow~\citet{zhang-etal-2023-macsum} to train the HP+SP baseline and use the final checkpoint. In all our experiments, we use beam search with beam size 3, and follow the default decoding settings of the Transformer implementation (temperature = 1.0, top\_k = 50, top\_p = 1.0). We adopt beam search since each node represents a full summary rather than a token- or sentence-level fragment. In this setting, greedy decoding can harm fluency and often generates lower-quality summaries, whereas a small beam (beam = 3) improves generation quality without introducing excessive branching. Moreover, the action space is fixed to five operations (corresponding to the number of attributes), and the maximum tree depth is also limited to five. As a result, the search tree remains very shallow.

\section{Hardware Usage}

We used 4 NVIDIA A100-SXM4-80GB GPUs for our experiments.

\section{Balancing Between Deterministic and Non-deterministic Attributes}
\label{appendix:balancing}
To evaluate each node during the search, we compute the local reward as $\frac{\alpha}{avg_{\text{det}} + \varepsilon} + \frac{1}{\beta} \cdot avg_{\text{non-det}}$, where $avg_{det}$ denotes the average deviation from the target values for \textit{deterministic} attributes, and $\text{avg}_{\text{non-det}}$ represents the average affinity score for \textit{non-deterministic} attributes. The hyperparameters $\alpha$ and $\beta$ control the relative importance of the two terms, and $\varepsilon$ is a small constant added to ensure numerical stability.
Since LLMs generally struggle more with structural, \textit{deterministic} attributes (e.g., \textit{length}, \textit{extractiveness}) than with content-related, \textit{non-deterministic} attributes, we set $\beta = 10$ in the main results (Tables~\ref{tab:main_dial} and \ref{tab:main_doc}) to mildly upweight \textit{deterministic} control performance.


In Table~\ref{tab:beta}, we vary the value of $\beta$ to adjust the relative weighting between \textit{deterministic} and \textit{non-deterministic} attributes in our experiments. We conduct experiments on the MACSum$_\text{Dial}$ dataset using Qwen2.5-7B and Llama-3.3-70B as baseline models. A smaller $\beta$ value increases the relative weight assigned to \textit{non-deterministic} attributes. The experimental results show that as $\beta$ decreases, the scores for \textit{non-deterministic} attributes such as \textit{topic} and \textit{speaker} gradually increase, while those for \textit{deterministic} attributes like \textit{extractiveness}, \textit{length}, and \textit{specificity} tend to decrease. 
Specifically, with the Qwen2.5-7B when $\beta = 10$, the MAD for length was 11.79, and the topic score was 0.799. However, when $\beta$ was reduced to 0.01 to emphasize \textit{non-deterministic} attributes, the MAD for length increased to 17.94, while the topic score improved to 0.801. A similar trend was observed with the Llama-3.3-70B model.
These findings suggest that users can control the summarization output by adjusting weights to emphasize the attributes they value most.

\begin{table}[t]
\centering
\scalebox{0.68}{
\begin{tabular}{l|c|c}
\toprule
\textbf{Model} & \textbf{\# of Params} & \textbf{Time (s) per Summary} \\
\midrule
Llama-3.3-Instruct      & 70B        & 22.826  \\
\quad$\circ$ Implicit self-planning  & 70B    & 33.410  \\
\quad$\circ$ Explicit self-planning  & 70B    & 104.312 \\
\quad$\circ$ Explicit self-planning+  & 70B   & 90.941  \\
\quad$\circ$ PACO            & 70B      & 196.380 \\
\bottomrule
\end{tabular}}
\caption{Average time required to generate a single summary.}
\label{tab:time_per_summary}
\end{table}

\section{Computational Costs}
\label{appendix:computational cost}

In Table~\ref{tab:time_per_summary}, we present the average time required by each model to generate a final summary. Although PACO incurs a higher computational cost, it clearly outperforms self-planning approaches, which are also relatively expensive. This demonstrates a favorable trade-off between computation and controllability, particularly in tasks that require structured control. Importantly, our method tackles the complex challenge of multi-attribute control entirely through test-time inference. Since higher computational cost is often necessary for stronger reasoning and controllability, and LMs are becoming faster and more efficient, we consider the increased computational cost required for stronger controllability to be practical and promising rather than a long-term limitation.


\section{Attribute Control Prompts}
\label{appendix:attribute control prompts}

The following are the detailed prompts used for attribute control. These prompts are shared across both PACO and the self-planning methods to ensure a fair comparison.

\subsection{Initial prompts}

\begin{quote}
{\ttfamily
You are a helpful assistant. Your task is to generate adjusted summary for user.
\\
\\
article: 
\\
\{\{Article\}\}
\\
\\
Summarize the above article in exactly \{\{length\}\} words focusing on \{\{topic\}\} and \{\{speaker\}\} while retaining \{\{extractiveness\}\} of the words verbatim from the article, and including \{\{specificity\}\} of the detailed information based on named entities. Ensure the summary is well-written, logically sound, with clear sentence flow.
\\
\\
summary (generate summary ONLY):
}\end{quote}

\subsection{Extractiveness control prompts}

\begin{quote}
{\ttfamily
You are a helpful assistant. Your task is to generate adjusted summary for user.
\\
\\
article: 
\\
\{\{Article\}\}
\\
\\
\{\{History\}\}
\\
\\
summary: 
\\
\{\{Previous summary\}\}
\\
\\
The summary you previously generated did not follow the given instructions. Summarize the above article while retaining \{\{extractiveness\}\} of the words verbatim from the article. Ensure the summary is well-written, logically sound, with clear sentence flow.
\\
\\
summary (generate summary ONLY):
}
\end{quote}

\subsection{Length control prompts}

\begin{quote}
{\ttfamily
You are a helpful assistant. Your task is to generate adjusted summary for user.
\\
\\
article: 
\\
\{\{Article\}\}
\\
\\
\{\{History\}\}
\\
\\
summary: 
\\
\{\{Previous summary\}\}
\\
\\
The summary you previously generated did not follow the given instructions. Summarize the above article in exactly \{\{length\}\} words. Ensure the summary is well-written, logically sound, with clear sentence flow.
\\
\\
summary (generate summary ONLY):
}
\end{quote}

\subsection{Specificity control prompts}

\begin{quote}
{\ttfamily
You are a helpful assistant. Your task is to generate adjusted summary for user.
\\
\\
article: 
\\
\{\{Article\}\}
\\
\\
\{\{History\}\}
\\
\\
summary: 
\\
\{\{Previous summary\}\}
\\
\\
The summary you previously generated did not follow the given instructions. Summarize the above article including \{\{specificity\}\} of the detailed information based on named entities. Ensure the summary is well-written, logically sound, with clear sentence flow.
\\
\\
summary (generate summary ONLY):
}
\end{quote}

\subsection{Topic control prompts}

\begin{quote}
{\ttfamily
You are a helpful assistant. Your task is to generate adjusted summary for user.
\\
\\
article: 
\\
\{\{Article\}\}
\\
\\
\{\{History\}\}
\\
\\
summary: 
\\
\{\{Previous summary\}\}
\\
\\
The summary you previously generated did not follow the given instructions. Summarize the above article focusing on topic \{\{topic\}\}. Ensure the summary is well-written, logically sound, with clear sentence flow.
\\
\\
summary (generate summary ONLY):
}
\end{quote}

\subsection{Speaker control prompts}

\begin{quote}
{\ttfamily
You are a helpful assistant. Your task is to generate adjusted summary for user.
\\
\\
article: 
\\
\{\{Article\}\}
\\
\\
\{\{History\}\}
\\
\\
summary: 
\\
\{\{Previous summary\}\}
\\
\\
The summary you previously generated did not follow the given instructions. Summarize the above article focusing on speaker \{\{speaker\}\}. Ensure the summary is well-written, logically sound, with clear sentence flow.
\\
\\
summary (generate summary ONLY):
}
\end{quote}

\section{LLM-Based Self-Planning}
\label{appendix:self-planning}

\subsection{Implicit self-planning}

\begin{quote}
{\ttfamily
You are a helpful assistant. Your task is to generate adjusted summary for user.
\\
\\
article: 
\\
\{\{Article\}\}
\\
\\
\{\{Initial prompts\}\}
\\
\\
summary: 
\\
\{\{Previous summary\}\}
\\
\\
You must modify \{\{target attributes\}\}, but since it is difficult to modify them all at once, you should adjust them one by one. Let's think step by step, internally consider the order in which to adjust the attributes, and gradually revise the summary to generate one that satisfies all attributes.
\\
\\
summary (generate summary ONLY):
}\end{quote}

\subsection{Explicit self-planning}

\begin{quote}
{\ttfamily
You are a helpful assistant. Your task is to generate adjusted summary for user.
\\
\\
article: 
\\
\{\{Article\}\}
\\
\\
\{\{Initial prompts\}\}
\\
\\
summary: 
\\
\{\{Previous summary\}\}
\\
\\
You are a helpful assistant. Your task is to generate a plan for adjusting summary.
\\
\\
You must modify \{\{target attributes\}\}, but since it is difficult to modify them all at once, you should adjust them one by one. Plan which attribute should be modified first. The output should be returned as a list. For example, plan = ['attribute1', 'attribute2', ...]
\\
\\
plan (generate plan ONLY):
}\end{quote}

\subsection{Explicit self-planning+}

\begin{quote}
{\ttfamily
You are a helpful assistant. Your task is to generate adjusted summary for user.
\\
\\
article: 
\\
\{\{Article\}\}
\\
\\
\{\{Initial prompts\}\}
\\
\\
summary: 
\\
\{\{Previous summary\}\}
\\
\\
You are a helpful assistant. Your task is to generate a plan for adjusting summary.
\\
\\
You must modify \{\{target attributes\}\}, but since it is difficult to modify them all at once, you should adjust them one by one. Plan which attribute should be modified first. Note that you do not need to modify all attributes, and you may adjust the same attribute multiple times if necessary. The output should be returned as a list. For example, plan = ['attribute1', 'attribute2', ...]
\\
\\
plan (generate plan ONLY):
}\end{quote}

\section{LLM-based Heuristic Score}
\label{appendix:heuristic}

\begin{quote}
{\ttfamily
You are a helpful assistant. Your task is to assess the summary for user.
\\
\\
article: 
\\
\{\{Article\}\}
\\
\\
summary:
\\
\{\{Summary\}\}
\\
\\
Can further adjustments to this summary fully satisfy all target attributes? The final objective is to generate a summary that meets the target attributes \{\{target attributes\}\}. The current summary has been adjusted in the order of \{\{path\}\}. Keep in mind that changes made to earlier attributes might be disrupted when you adjust later ones.
\\
\\
answer (generate Yes or No ONLY):
}\end{quote}

\end{document}